\RequirePackage{snapshot}
\documentclass{article}
\usepackage{iclr2015,times}

\pdfoutput=1

\usepackage[utf8]{inputenc}

\usepackage{amsmath}
\usepackage{amsfonts}
\usepackage{dsfont}
\usepackage{amssymb}
\usepackage[per-mode = symbol]{siunitx}
\usepackage{hyperref}
\usepackage{graphicx}
\usepackage{todonotes}

\usepackage{natbib}
\newcommand{\fix}{\marginpar{FIX}}

\iclrfinalcopy

\newcommand*{\noktikz}{}
\usepackage{tikz}
\usetikzlibrary{calc}
\usetikzlibrary{shapes}

\graphicspath{{figures/}}

\DeclareSIUnit\fix{fix}
\DeclareSIUnit\bits{bits}
\DeclareSIUnit\bit{bit}

\expandafter\xdef\csname factorBestToDeepSal\endcsname{1.7}
\expandafter\xdef\csname factorBesteToDeepSal\endcsname{1.00}
\expandafter\xdef\csname percentIgBest\endcsname{34}
\expandafter\xdef\csname factorBestToDeepSalPercent\endcsname{67}
\expandafter\xdef\csname percentBaseline\endcsname{42}
\expandafter\xdef\csname percentIgDeepSal\endcsname{56}

\title{Deep Gaze I: Boosting saliency prediction with feature maps trained on ImageNet}

\author{Matthias Kümmerer, Lucas Theis \& Matthias Bethge \\
Werner Reichardt Centre for Integrative Neuroscience \\
University Tübingen, Germany \\
\texttt{\{matthias.kuemmerer,lucas,matthias\}@bethgelab.org}
}

\begin{document}

\maketitle

\begin{abstract}
  \noindent Recent results suggest that state-of-the-art saliency models perform far from optimal
  in predicting fixations. This lack in performance has been attributed to an inability to model the influence of
  high-level image features such as objects. Recent seminal advances in
  applying deep neural networks to tasks like object recognition suggests that they are able
  to capture this kind of structure. However, the enormous amount of training data necessary
  to train these networks makes them difficult to apply directly to saliency prediction.
  We present a novel way of reusing existing neural networks that have been pretrained on the
  task of object recognition in models of fixation prediction.
  Using the well-known network of \cite{krizhevsky2012},
  we come up with a new saliency model that significantly outperforms
  all state-of-the-art models on the MIT Saliency Benchmark.
  The structure of this network allows new insights in the
  psychophysics of fixation selection and potentially their neural implementation. To train our network,
  we build on recent work on the modeling of saliency as point processes.
\end{abstract}

\noindent
By understanding how humans choose eye fixations, we can hope to understand and explain human behaviour
in a number of vision-related tasks. For this reason human eye movements have been studied for more
than 80 years \citep[e.\,g. ][]{Buswell1935people}. During the last 20 years,
many models have been developed trying to explain fixations in terms of so called ``saliency maps''.



Recently, it has been suggested to model saliency maps probabilistically using point processes \citep{Barthelme} and to evaluate them using log-likelihood \citep{Kuemmerer2014}.
This evaluation revealed that state-of-the-art models of saliency explain only one third of the explainable information in the spatial fixation structure
\citep{Kuemmerer2014}.

\begin{figure}[t]
  \begin{center}
  \input{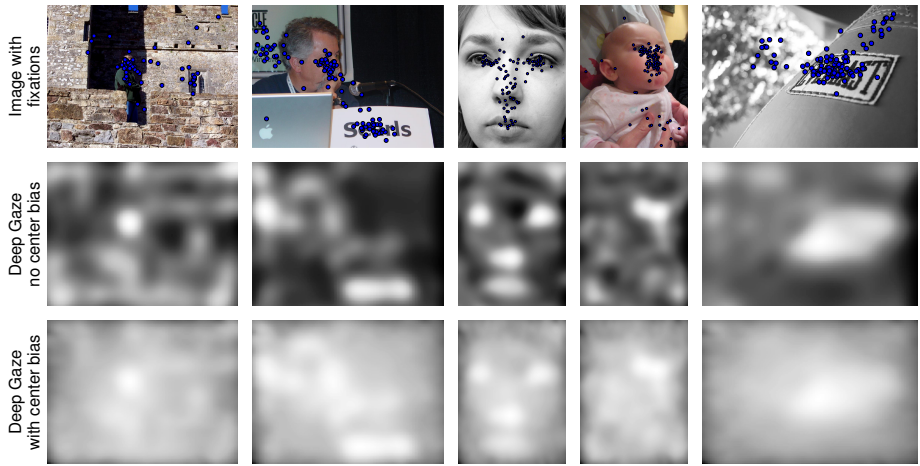}
  \end{center}

  \caption{
    Example saliency maps: The top row shows example images from the dataset by \cite{Judd2009Model}.
    The fixations of the subjects are indicated by dots.
    The middle row shows the log-densities produced by Deep Gaze I for these images when assuming a uniform prior distribution instead of a center bias.
    The bottom row shows the log-densities for the same images when using the center bias of the full dataset.
    Note that only the first two images were included in the set of images used to train Deep Gaze I.}
  \label{fig:saliency_maps}
\end{figure}

Most of the existing models use low-level cues like edge-detectors and color filters \citep{IttiKoch1998Model} or local image statistics \citep{Zhang2008SUN,BruceTsotso2009Saliency}.
However, human fixations are largely clustered around objects (see Figure \ref{fig:saliency_maps} for examples).
This has led to some models trying to incorporate more high level features:
\cite{cerf2007} combined existing saliency map models with a face detector, while \cite{Judd2009Model} included detectors for faces, people, cars and horizon.
Nevertheless, current saliency models mostly fail to capture these high-level influences which might be the main reason for the poor overall performance of state-of-the-art models.
This analysis raises the question whether there are any computational systems capable of capturing such high-level influences.

Independent of these developments, the last two years have seen the rise of deep neural networks to solve multifarious tasks like object detection, speech recognition or automatic translation.
Provided with enough training data, deep neural networks show impressive results, often outperforming all competing methods.
It has also been shown that deep convolutional networks that have been optimized for object
classification can be used to predict neuron responses in higher brain areas of the visual system \citep{yamins2014performance,razavian2014cnn}.
Deep neural networks have also proven to generalize well over tasks \citep{donahue2013decaf}:
a network trained for some task like object detection can often be easily retrained to achieve
state-of-the-art performance in some other only loosely related task like scene recognition.

Motivated by these developments, we here try to use pretrained deep neural networks to model fixation selection.
The results of \cite{yamins2014performance} connect neural network representations with IT and
similar neural representations. This suggests that we can hope not only to improve prediction performance,
but also to improve our understanding of the internal implementation of fixation selection in the
brain by formulating new hypotheses that lead to new experimental paradigms.
Finally, results from \cite{Zeiler2013visualizing} show ways to interpret the filters of deeper
layers in a way that would allow to formulate predictions that can be tested psychophysically.

A first attempt at modelling saliency with deep convolutional networks has been performed recently by \cite{Vig2014} (eDN), yielding state-of-the-art performance.
However, training deep neural networks on fixations suffers from the usually small training sets
compared to the training data used in other tasks. To reach their state-of-the-art performance,
neural networks trained for object or speech recognition need massive amounts of training data.
Most fixation datasets have at most 1000 images with usually not significantly more than 100 fixations per image.
Deep neural networks can easily have millions of parameters, which would lead to massive overfitting on these small datasets.
Therefore, eDN uses only three convolutional layers, while the Krizhevsky network uses 5
convolutional layers and the most recent networks used in the ImageNet challenge (ILSVRC2014) use
around 20 layers.

Here we present a new model of fixation prediction that builds on these results:
it uses the well known deep network from \cite{krizhevsky2012} to generate a high-dimensional
feature space, which is then used for the actual fixation prediction. This deep network has been
optimized for object recognition using a massive dataset consisting of more than one million images
\citep{deng2009imagenet}.
Keeping the parameters of the deep network fixed, we train
our model on half of the MIT1003 dataset \citep{Judd2009Model} and show that it outperforms state-of-the-art
models by a large margin, increasing the amount of explained information by
\SI{\factorBestToDeepSalPercent}{\percent}. Furthermore, we analyze
how the model exploited the feature space provided by the Krizhevsky network.

\begin{figure}[t]
  \center
  \input{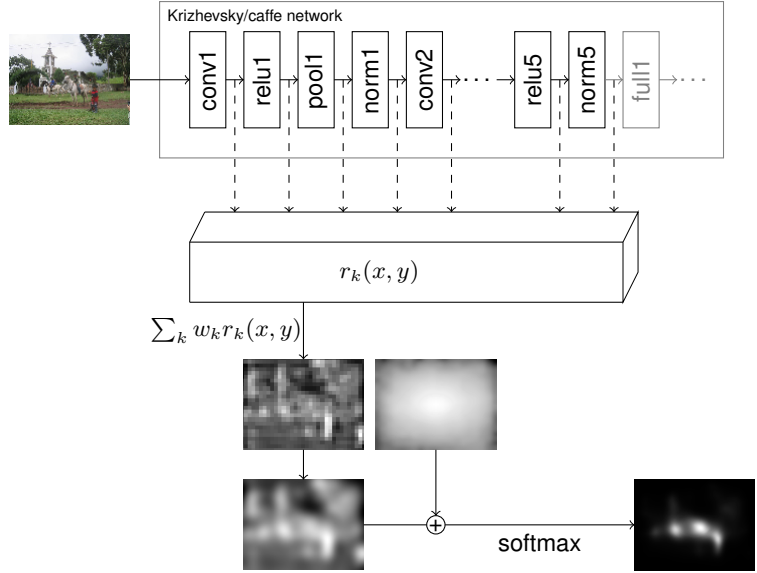}
  \caption{The model structure of Deep Gaze I: The image is first downsampled and preprocessed with the Krizhevsky network.
    The responses of the layers that are included in the model are then scaled up to the size of the
    largest network layer and normalized to have unit standard deviation.
    This list of maps is then linearly combined and blured with a Gaussian kernel.
    To compensate for the central fixation bias, an estimate of the prior distribution is added.
    Finally, the model output is fed through a softmax rectification, yielding a two dimensional probability distribution.
    }
    \label{fig:architecture}
\end{figure}



\section{Methods}
\label{sec:methods}

In \autoref{fig:architecture}, the model architecture is visualized.
After an initial downsampling, the RGB input image is fed into the Krizhevsky network.
The Krizhevsky architecture consists of stacked convolutions, each one followed by a rectifiying nonlinearity and optional maxpooling and response normalization.
The final three fully connected layers of the Krizhevsky network were removed as we are only interested in spatially located features.
Each layer (convolution, rectifier, pooling and normalization) results in a single image of response for each filter in the layer.
To predict fixations, we first select one or multiple layers from the network.
We rescale all the response images that we want to include in our model to the size of the largest
layer of the network, resulting in a list of up to 3712 responses for each location in an image.
Each of these responses is then individually normalized to have unit standard deviation on the full dataset. After
this preprocessing, the features are fed into the following model.

At leach image location, our saliency model linearly combines the responses $r_k(x, y)$ using weights $w_k$.
The resulting image is then convoled with a Gaussian kernel whose width is controlled by $\sigma$,
yielding the saliency map
\[
  s(x, y) = \sum_k w_k r_k(x, y) * G_\sigma.
\]

\noindent It is well known that fixation locations are strongly biased towards the center of an image \citep{Tatler2007Centerbias}. 
To account for this center bias, the saliency prediction is linearly combined with a fixed center bias prediction $c(x, y)$:

\[
  o(x, y) = \alpha c(x, y) + s(x, y)
\]

\noindent To predict fixation probabilities, this output is finally fed into a softmax, yielding a probability distribution over the image:

\[
    p(x, y) = \frac{\exp\left(o(x, y)\right)}{\sum_{x,y} \exp\left(o(x,y)\right)}
\]

For generalization, $\ell_1$-regularization on the weights is used to encourage sparsity. For
training fixations $(x_1, y_1), \dots, (x_N, y_N)$ this yields the cost function


\[
  c(\mu, \alpha, w) = -\frac{1}{N} \sum_{i}^N \log p(x_i, y_i) + \lambda \frac{\|w\|_1}{\|w\|_2}
\]

To quantify which layers help most in predicting the fixations and lead to least overfitting, we trained models on a variety of subsets of layers (see \autoref{sec:selecting_layers} and \autoref{fig:layer_restriction_results}).
We checked the generalization performance of these models on the remaining 540 images from MIT1003 that have not been used in training.
As performance measure we use shuffled area under the curve (shuffled AUC) here \citep{Tatler2005ROC}.
In AUC, the saliency map is treated as a classifier score to separate fixations from ``nonfixations'': 
presented with two locations in the image, the classifier chooses the location with the higher saliency value as fixation.
The AUC measures the classification performance of this classifer.
The standard AUC uses a uniform nonfixation distribution, while in the case of shuffled AUC, fixations from other images are used as nonfixations.
As shuffled AUC assumes the saliency maps not include the biases of the prior distribution \citep[see][]{Barthelme} we had to use a uniform center bias for this evaluation.


\subsection{Implementation details}
\label{sec:implementation_details}

For training, we used roughly half of the dataset MIT1003 \citep{Judd2009Model}.
By using only the images of the most common size of $1024\times 768$ pixels (resulting in 463 images), we were able to use the nonparametric estimate of the center bias described in \cite{Kuemmerer2014} (mainly a 2d histrogram distribution fitted using the fixations from all other images).

Our implementation of the Krizhevsky network uses the architecture and trained filters as published by \cite{jia2014caffe} with the following modifications:
the original architecture uses a fixed input size of $224\times 224$.
As we removed the fully connected layers, we do not need to restrict to a fixed input size but can feed arbitrary images into the network.
Furthermore we use convolutions of type \textit{full} (i.e. zero-pad the input) instead of \textit{valid} which would result in convolution outputs that are smaller than the input.
This modification is useful, because we need saliency predictions for every point in the image.
Note that the caffe implementation of the Krizhevsky network differs slightly from the original architecture in \cite{krizhevsky2012}, as the pooling and the normalization layers have been switched.
The subsampling factor for the inital downsampling of the images was set to 2.

The sparsity parameter $\lambda$ was chosen using grid search and turned out to be $0.001$ in the final model.
However, even setting it to much smaller values did have very little effect on training and test performance (see \autoref{sec:regularizaton} for more details).
All calculations of log-likelihoods, cost functions and gradients were done in theano \citep{bergstra+al:2010-scipy}.
To minimize the cost function on the training set of fixations, the mini-batch based BFGS method as described in \cite{Sohl-Dickstein2013} was used.
It combines the benefits of batch based methods with the advantage of second order methods, yielding high convergence rates with next to no hyperparameter tuning.
To avoid overfitting to the subjects, leave-one-out cross-validation over the 15 subjects contained in the database was used.

The code for our model including training and analysis will be published at
\url{http://www.bethgelab.org/code/deepgaze/}.

\section{Results}
\label{sec:results}

\subsection{Performance Results}
\label{sec:performance_results}

\begin{figure}
  \begin{center}
    \input{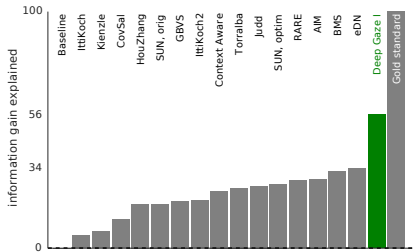}
  \end{center}

  \caption{
  Performance of Deep Gaze I compared to a list of other influential models, expressed as the ratio of explained information (see text for details).
  All models except for Deep Gaze I have been postprocessed to account for a pointwise nonlinearity, center bias and blurring (see \cite{Kuemmerer2014} for details).
  }
  \label{fig:loglikelihoods}
\end{figure}

We use an information theoretic measure to evaluate our model: log-likelihood.
Log-likelihood is a principled measure for probabilistic models and has numerous advantages. See \cite{Kuemmerer2014} for an extensive discussion.

Log-likelihoods are much easier to understand when expressed as difference of log-likelihood relative to a baseline model.
This \textit{information gain}\footnote{To be more precise, this value is an estimated expected information gain} expresses how much more efficient the model is in describing the fixations than the baseline model:
if a model with an information gain of \SI{1}{\bit\per\fix} is used to encode fixation data, it can save on average one bit per fixation compared to the baseline model.

The information gain is even more intuitive when compared to the explainable information gain, i.e., the information gain of the real distribution compared to the baseline model.
This comparison yields a ratio of explained information gain to explainable information gain which
will be called ``explainable information gain explained'' or just ``information gain explained'' in
the following.
See \cite{Kuemmerer2014} for a more thorough explanation of this notion.

The baseline model is a non-parametric model of the image-independent prior distribution $p(x,y)$,
while the explainable information is estimated using a non-parametric model of the fixation
distribution $p(x, y \mid I)$ for a given image $I$ (which we call the \textit{gold standard model}).
The gold standard model is cross-validated between subjects and thus captures all the structure in the fixations that is
purely due to the spatial structure of the image.
See \cite{Kuemmerer2014} for details on the baseline model and the gold standard model.

By expressing the information gain of a model as a percentage of the possible information gain, we can asses how far we have come in describing the fixations.
It is important to note that this interpretation is only possible due to the fact that information gain is on a ratio scale \citep{michell1997quantitative}:
differences and ratios of information gains are meaningful -- opposed to other measures like AUC.

In \autoref{fig:loglikelihoods}, the percentage of information gain explained is plotted for our model in comparison to a range of influential saliency models, including the state-of-the-art models.
Of the possible information gain, the best existing model (eDN) is able to explain only \SI{\percentIgBest}{\percent}.
Deep Gaze I is able to increase this information gain to \SI{\percentIgDeepSal}{\percent}.

\begin{figure}[ht!]
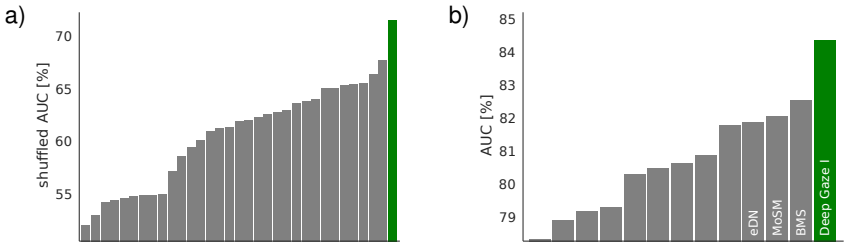

\begin{center}
  \include{figures/mit_results.pgf}
\end{center}
  \caption{Performance results on the MIT benchmark.
    \textbf{(a)}: Shuffled AUC performance of Deep Gaze I (green bar, 71.69\%) compared with all other models in the MIT benchmark.
    The x-axis is at the level of the center bias model.
    The three top performing models after Deep Gaze I are in order of decreasing performance:
    AWS (67.90\%, \cite{garcia2012relationship}), 
    RARE2012 (66.54\%, \cite{Riche2013a}), and
    AIM (65.64\%, \cite{BruceTsotso2009Saliency}).
    \textbf{(b)} AUC performance of Deep Gaze I (green bar, 84.40\%) compared with all other models in the MIT benchmark that performed better than the center bias.
    The x-axis is at the level of the center bias model.
    The three top performing models after Deep Gaze I are in order of decreasing performance:
    BMS (82.57\%, \cite{zhang2013saliency}),
    Mixture of Saliency Models (82.09\%, Han and Satoh, 2014), and
    eDN (81.92\%, \cite{Vig2014}).
    Notice that AUC and shuffled AUC use different definitions of saliency map:
    While AUC expects the saliency maps to model the center bias,
    shuffled AUC explicitly does not and penalizes models that do.
    Therefore, for the shuffled AUC performances of Deep Gaze I the saliency maps have been calculated with a uniform prior distribution,
    while for the AUC performances the saliency maps have been calculated with a nonparametric prior (see text for details)
    \protect\footnotemark[2]. 
    Performances of other models from the MIT benchmark as of
    September 2014.
    }
    \label{fig:mit-benchmark}
\end{figure}

\begin{figure}[ht!]
  \begin{center}
    \input{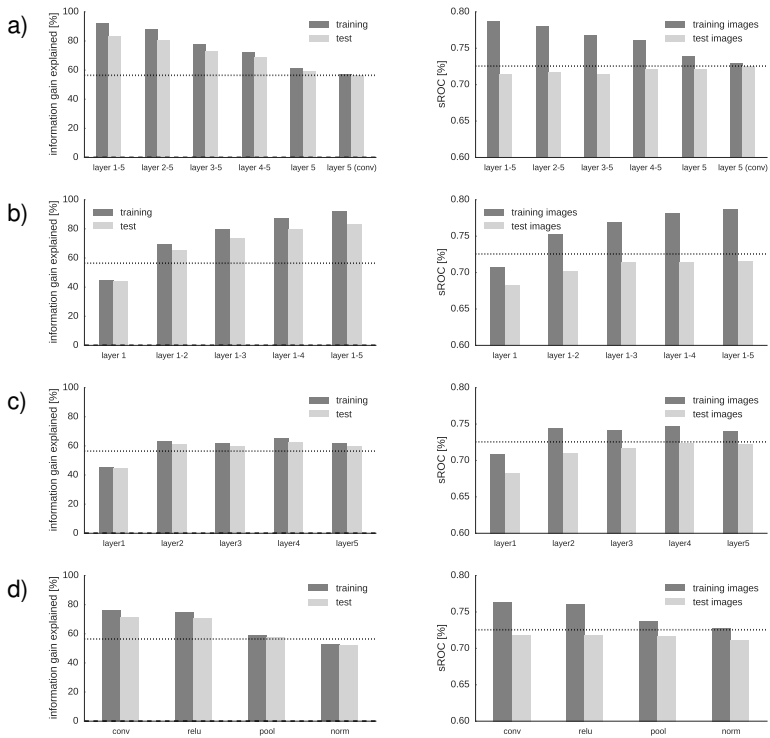}
  \end{center}
  \caption{Performance of Deep Gaze I when trained on different subsets of the Krizhevsky layers:
    \textbf{(a)}: Results for models that use layers from a given depth upwards.
    The left plot shows the percentage of explainable information gain explained on the images used in training for training subjects and test subjects
    (refer to \autoref{sec:performance_results} for an explanation of this measure).
    The dotted line indicates the performance of the model we used in the MIT Saliency Benchmark
    (which only used the output of the convolutions of layer 5).
    The right plot shows the shuffled AUC on the images used in training and on the remaining test images.
    Here, the models have been averaged over all test subjects and the saliency maps assume uniform center bias,
    as expected by shuffled AUC (see \autoref{sec:mit-benchmark} for details).
    The dotted line indicates the performance of the final model on the test images.
    \textbf{(b), (c), (d)}: Results for models that use layers up to a given depth (b),
    layers of a certain depth (c) and layers of a certain type (d).
    The plots are as in (a).}
  \label{fig:layer_restriction_results}
\end{figure}

\begin{figure}[h!]
  \begin{center}
    \input{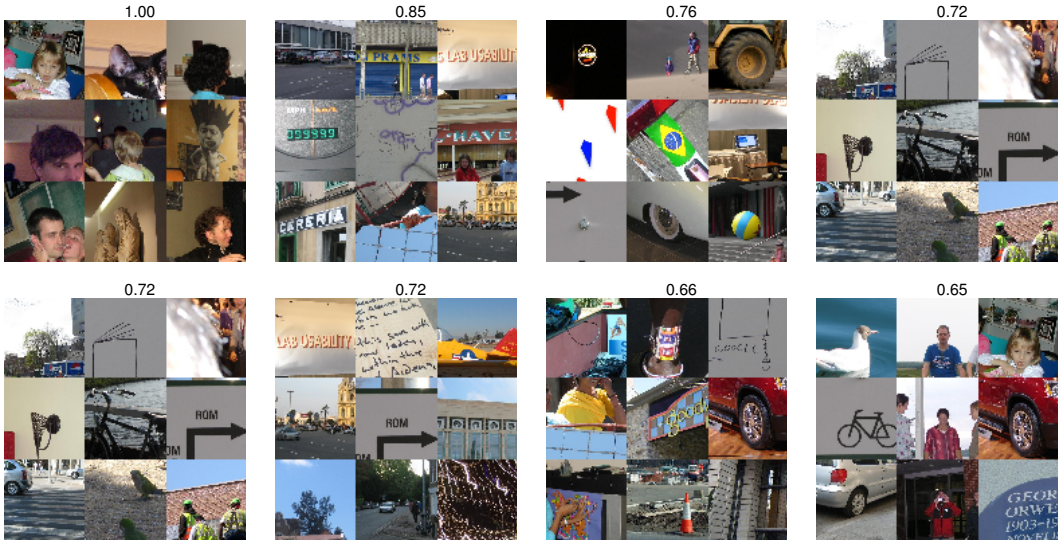}
  \end{center}

  \caption{Analysis of used features I:
    \textbf{(a)} Patches of maximum response:
    Each square of patches shows for a specific feature of the Krizhevsky architecture the nine patches that led to highest response (resp.~smallest response, if the feature has a negative weight in the model).
    Each patch corresponds to exactly the part of the image that contributes to the response in the location of maximum response.
    The features used have been choosen by the absolute value of the weight that Deep Gaze I assigned to them.
    The numbers over the patches show $|w_k|/\max_k |w_k|$.
    }
  \label{fig:used_features}
\end{figure}

\begin{figure}[ht!]
  \begin{center}
    \input{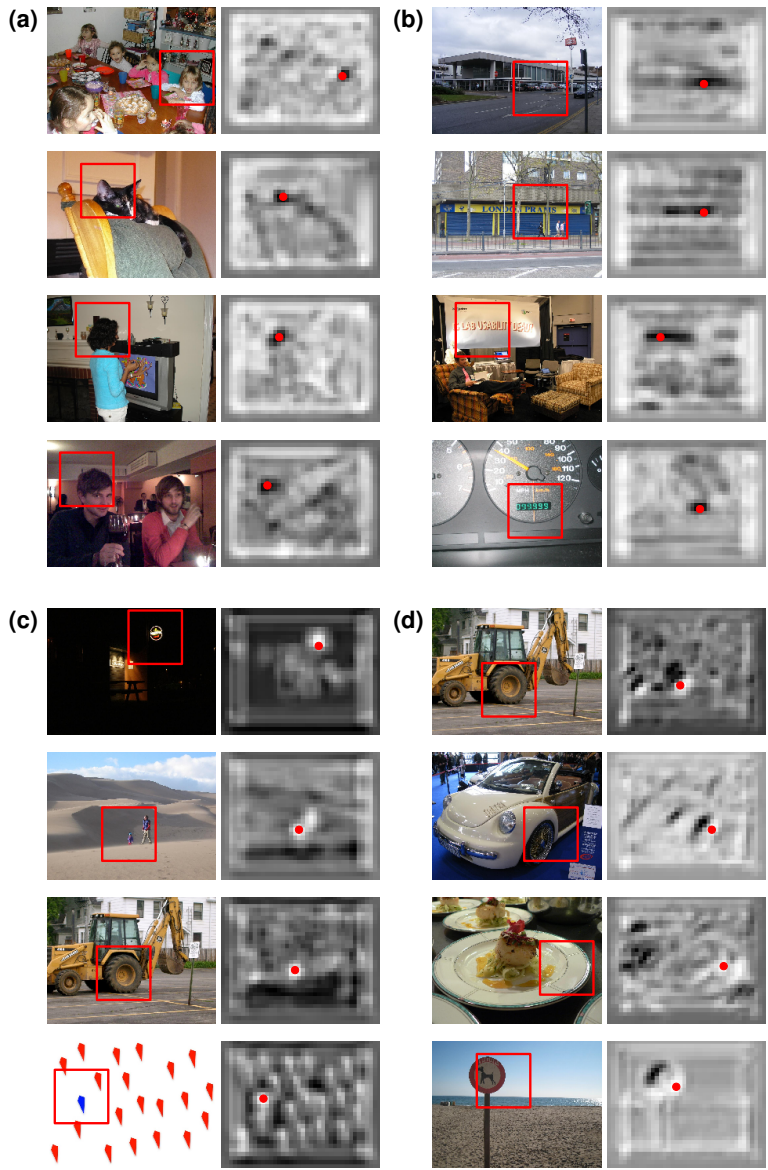}
  \end{center}

  \caption{Analysis of used features II:
    Details for some of the patches from \autoref{fig:used_features}
    The four double columns (a) to (d) correspond to the first four features shown \autoref{fig:used_features}.
    In each double column, the four rows correspond to the first four patches shown for this feature in \autoref{fig:used_features}.
    The left column of each double column shows the patches in the context of the full image,
    while the feature's response over the full image is shown in the right column.
    The position of the maximum is indicated by a dot.
    }
  \label{fig:used_features_context}
\end{figure}

\subsection{Results on MIT Saliency Benchmark}
\label{sec:mit-benchmark}

We submitted our model to the MIT Saliency Benchmark (\cite{mit-saliency-benchmark}).
The benchmark evaluates saliency models on a dataset of 300 images and 40 subjects.
The fixations are not available to make training for these fixations impossible.

The MIT Saliency Benchmark evaluates models on a variety of metrics, including AUC with uniform nonfixation distribution and shuffled AUC (i.e. AUC with center bias as nonfixation distribution).
The problem with these metrics is that most of them use different definitions of saliency maps.
This hold especially for the two most used performance metrics: AUC and shuffled AUC.
While AUC expects the saliency maps to model the center bias,
shuffled AUC explicitly does not so and penalizes models that do
(see \cite{Barthelme} for details).
As Deep Gaze I uses an explicit representation of the prior distribution, it is straightforward to produce the saliency maps according to both definitions of 
AUC: For AUC we use a nonparametric prior estimate, for shuffled AUC we use a uniform prior distribution.
As the images of the dataset are of different size, we could not use our non-parametric center bias as is.
Instead, we took all fixations from the full MIT-1003 dataset and transformed their position to be relative to a image of size $100\times 100$. Then we trained a Gaussian kernel density estimator on these fixations. This density estimate was then rescaled and renormalized for each image.

Doing so, we beat the state-of-the-art models in the MIT Saliency Benchmark by a large margin in AUC as well as shuffled AUC (see \autoref{fig:mit-benchmark}):
For shuffled AUC, we reach 71.69\% compared to 67.90\% for the best performing model AWS (center bias is at 50\%).
For AUC we reach 84.40\% compared to 82.57\% for the best performing model BMS (center bias is at 78.31\%).
Relative to the center bias, this is an increase of AUC performance by more than 40\%.

\footnotetext[2]{Note that the MIT Saliency Benchmark webpage reports only performances for the saliency maps with the nonparametric prior.
Therefore, there the shuffled AUC performance is lower.}

\subsection{Layer selection}
\label{sec:selecting_layers}

The final model used only the convolutions of the top-most layer of the Krizhevsky-architecture.
This is a principled choice:
the top layer can be expected to include most high-level influences and the relu, pool and norm units are often viewed mainly as the nonlinearities needed to provide a new feature space for the next level of convolutions.

But this choice was also backed by a series of comparison models where more or other layers have been included in the model:
In \autoref{fig:layer_restriction_results}, performance results are reported for models including layers from a given depth upwards  (\autoref{fig:layer_restriction_results}a), layers up to a given depth (\autoref{fig:layer_restriction_results}b), layers of a given depth  (\autoref{fig:layer_restriction_results}c) and layers of a given type  (\autoref{fig:layer_restriction_results}d).
It can be seen that the architecture chosen finally (layer 5 convolutions) generalizes best to the images of the test set in terms of shuffled AUC.

It is also worth noting that models including more layers are substantially better at predicting the test subjects fixations on the images used in training (\autoref{fig:layer_restriction_results}a, left plot): when using all layers, a performance of 83\% information gain explained is reached for the test subjects.
This suggests that the generalization problems of these models are not due to intersubject variability. They most probably suffer from the fact that the variety of objects in the training images is not rich enough, leading to overfitting to the images (not to the subjects).
Therefore we can expect improved performance from using a larger set of images in training.

\subsection{Analysis of used features}
\label{sec:used_features}

In this section we analyze which features of the Krizhevsky architecture contributed most to the
fixation predictions.
By getting a solid understanding of the involved features, we can hope to extract predictions from
the model that can be tested psychophysically in the future.

In \autoref{fig:used_features}, we took the 10 most weighted features from the 256 convolution features in layer 5.
For each of these 10 features, we plotted the 9 patches from the dataset that led to the highest response
(resp. lowest response for features with negative weight).
In \autoref{fig:used_features_context}, the first four patches of the first four features are shown in more detail: The patches are shown in the context of the entire image and also the feature's response to this image is shown.

Clearly, the most important feature is sensitive to faces.
The second most important feature seems to respond mainly to text.
The third most important feature shows some sort of pop-out response:
it seems to respond to whichever feature sticks out from an image: the sign of a bar in the first
patch, two persons in a desert in the second patch and, most notably, the target in a visual search image in the fourth patch.
Note that the salient feature depends heavily on the image context, so that a simple luminance or color contrast detector would not achieve the same effect.

This shows that Deep Gaze I is not only able to capture the influence of high level objects like faces
or text, but also more abstract high-level concepts (like popout).


\section{Discussion}


Deep Gaze I was able to increase the explained information gain to \SI{\percentIgDeepSal}{\percent}
compared to \SI{\percentIgBest}{\percent} for state of the art models. On the
MIT Saliency Benchmark we were also able to beat the state of the art models by a substantial
margin.
One main reason for this performance is the ability of our model to capture the influence of several high-level features like faces and text but also more abstract ones like popout (\ref{sec:used_features}).

It is important to note that all reported results from Deep Gaze I are direct model performances, without any fitting of a pointwise nonlinearity as performed in \cite{Kuemmerer2014}.
This indicates that the deep layers provide a sufficiently rich feature space to enable fixation prediction via simple linear combination of the features.
The convolution responses turned out to be most informative about the fixations.

While features trained on ImageNet have been shown to generalize to other recognition and detection tasks \citep[e.\;g. ][]{donahue2013decaf, razavian2014cnn}, to our knowledge this is the first work where ImageNet features have been used to predict behaviour.

Extending state-of-the-art neural networks with attention is an exciting new direction of research
\citep{tang2014learning,mnih2014recurrent}.
Humans use attention for efficient object recognition and we showed that Krizhevsky features work well for predicting human attention.
Therefore it is likely that these networks could be brought closer to human performance by extending them with Krizhevsky features.
This could be an interesting field for future research.

\section{Conclusions}

Our contribution in this work is twofold:
First, we have shown that deep convolutional networks that have been trained on computer vision tasks like object detection boost saliency prediction.
Using the well-known Krizhevsky network \citep{krizhevsky2012}, we were able to outperform state-of-the-art saliency models by a large margin, increasing the amount of explained information by \SI{\factorBestToDeepSalPercent}{\percent} compared to state-of-the art.
We believe this approach will enable the creation of a new generation of saliency models with high predictive power and deep implications for psychophysics and neuroscience \citep{yamins2014performance,Zeiler2013visualizing}.
An obvious next step suggested by this approach is to replace the Krizhevsky network by the ImageNet 2014 winning networks such as VGG \citep{Simonyan2014} and GoogLeNet \citep{Szegedy2014}.

A second conceptual contribution of this work is to optimize the saliency model by maximizing the log-likelihood of a point process \citep[see ][]{Barthelme,Kuemmerer2014}.

We believe that the combination of high performance feature spaces for object recognition as obtained from the ImageNet benchmark with principled maximum likelihood learning opens the door for a ``Deep Gaze'' program towards explaining all the explainable information in the spatial image-based fixation structure.

\section{Acknowledgements}

This work was mainly supported by the German Research Foundation
(DFG; priority program 1527, Sachbeihilfe BE 3848-1) and additionally by
the German Ministry of Education, Science, Research and Technology through the Bernstein Center for
Computational Neuroscience (FKZ 01GQ1002) and the German Excellency Initiative through the Centre for Integrative Neuroscience Tübingen (EXC307).

\bibliography{literature}

\section{Supplementary Material}

\begin{figure}[ht!]
  \begin{center}
    \input{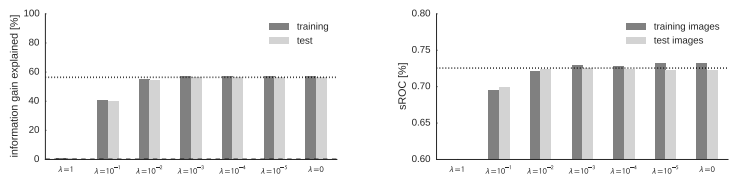}
  \end{center}
  \caption{Performance of Deep Gaze I when trained on the conv5-layer with different regularization parameters. 
    The left plot shows the percentage of explainable information gain explained on the images used in training for training subjects and test subjects
    (refer to \autoref{sec:performance_results} for an explanation of this measure).
    The dotted line indicates the performance of the model we used in the MIT Saliency Benchmark
    ($\lambda=0.001$).
    The right plot shows the shuffled AUC on the images used in training and on the remaining test images.
    Here, the models have been averaged over all test subjects and the saliency maps assume uniform center bias,
    as expected by shuffled AUC (see \autoref{sec:mit-benchmark} for details).
    The dotted line indicates the performance of the final model on the test images.
    }
  \label{fig:regularization_results}
\end{figure}

\subsection{Regularization}
\label{sec:regularizaton}

The model uses a regularization parameter $\lambda$ to encourage sparsity in the feature weights (see \autoref{sec:methods}).
This parameter was choosen using grid search.
In \autoref{fig:regularization_results}, training and test performances are shown for different choices of $\lambda$ when fitting the model using only the final convolutional layer (as done in the final model).
It can be seen that the choice of the regularization parameter had a visible but only very small effect on the test performance (especially if compared to the influences of the different layers used, see \autoref{fig:layer_restriction_results}).


\end{document}